\documentclass[11pt,a4paper]{article}
\usepackage[hyperref]{emnlp2020}
\usepackage{times}
\usepackage{latexsym}

\usepackage{graphicx} 
\graphicspath{ {images/} } 
\usepackage{multirow} 
\usepackage{bbding}
\usepackage{url}
\usepackage{enumitem}
\usepackage[normalem]{ulem}
\usepackage{xspace}

\usepackage{microtype}

\aclfinalcopy 
\def\aclpaperid{37} 

\title{\model: Improving Generalization of Clinical Text\\ De-identification Models via Data Augmentation}

\author{Xiang Yue \\
  The Ohio State University \\
  \texttt{yue.149@osu.edu} \\\And
  Shuang Zhou \\
  The Hong Kong Polytechnic University \\
  \texttt{shuang.zhou@connect.polyu.hk} \\}

\date{}

\newcommand{\model}{\textsc{PhiCon}\xspace}

\begin{document}
\maketitle
\setlist[itemize]{leftmargin=*}

\begin{abstract}
De-identification is the task of identifying protected health information (PHI) in the clinical text. Existing neural de-identification models often fail to generalize to a new dataset. We propose a simple yet effective data augmentation method \model to alleviate the generalization issue. \model consists of \textbf{PHI} augmentation and \textbf{Con}text augmentation, which creates augmented training corpora by replacing PHI entities with named-entities sampled from external sources, and by changing background context with synonym replacement or random word insertion, respectively. Experimental results on the i2b2 2006 and 2014 de-identification challenge datasets show that \model can help three selected de-identification models boost F1-score (by at most 8.6\%) on cross-dataset test. We also discuss how much augmentation to use and how each augmentation method influences the performance.\footnote{Our code is available at: \url{https://github.com/betterzhou/PHICON}}

\end{abstract}

\section{Introduction}
Clinical text in electronic health records (EHRs) often contain sensitive information. In the United States, Health Insurance Portability and Accountability Act (HIPPA)\footnote{\url{http://www.hhs.gov/hipaa}} requires that protected health information (PHI) (e.g., name, street address, phone number) must be removed before EHRs are shared for secondary uses such as clinical research \cite{meystre2014text}.

\begin{figure}[t]
    \centering
    \includegraphics[width=0.9\linewidth]{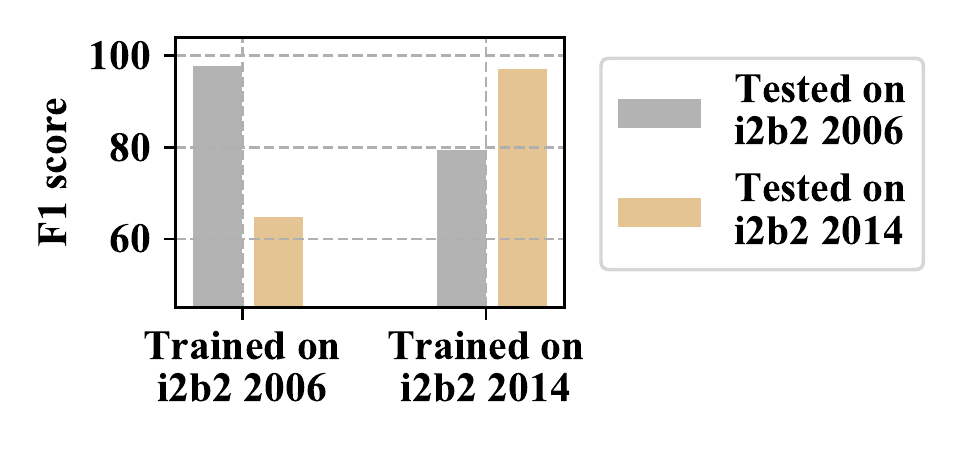}
    \caption{The result of cross-dataset test based on a base model \cite{Dernoncourt2017DeidentificationOP}. Performance on the new dataset drops up to 33\% compared to the original test set, showing the model suffers from generalizability issue.}
    \label{fig:de-id-crosstest}
\end{figure}

The task of identifying and removing PHI from clinical texts is referred as de-identification. Although many neural de-idenfication models such as LSTM-based \cite{Dernoncourt2017DeidentificationOP, Liu2017DeidentificationOC, Jiang2017DeidentificationOM, Khin2018ADL} and BERT-based \cite{alsentzer-etal-2019-publicly, Tang2019DeidentificationOC} have achieved very promising performance, identifying PHI still remains challenging in the real-world scenario: even well-trained models often \textit{fail to generalize to a new dataset}.
For example, we conduct cross-dataset test on i2b2 2006 and i2b2 2014 de-identification challenge datasets\footnote{\url{https://portal.dbmi.hms.harvard.edu/projects/n2c2-nlp/}} (i.e., train a widely-used de-identification model NeuroNER \cite{Dernoncourt2017DeidentificationOP} on one dataset and test it on 
the other one). The result in Figure \ref{fig:de-id-crosstest} shows that model's performance (F1-score) on the new dataset decreases up to $33\%$  compared to the original test set. The poor generalization issue on de-identification is also reported in previous studies \citep{Stubbs2017DeidentificationOP, Xi2019A, DBLP:conf/chil/JohnsonBP20, Tzvika2020Customization}. 

To explore what factors lead to poor generalization, we sample some error examples and find that the model might focus too much on specific entities and does not really learn language patterns well. For example, in Figure \ref{citation-figure2}, given a sentence \textit{``She met Washington in the Ohio Hospital"}, the model tends to recognize the entity \textit{``Washington"} as the ``Location" instead of the ``Name" if \textit{``Washington"} appears as ``Location" in the training many times. Such cases appear more frequently in a new testing set, thus leading to poor generalization.

To prevent the model overfitting on specific cases and encourage it to learn general language patterns, one possible way is to enlarge training data \cite{Xi2019A}. However, clinical texts are usually difficult to obtain, not to mention the requirement of tremendous expert effort for annotations \cite{yue2020CliniRC}. To solve this, we introduce our data augmentation method \model, which consists of \textbf{PHI} augmentation and \textbf{Con}text augmentation. Specifically, PHI augmentation replaces the original PHI entity in the training set with a same type named-entity sampled from external sources (such as Wikipedia). For example, in Figure \ref{citation-figure2}, \textit{``Ohio Hospital"} is replaced by an randomly-sampled ``Hospital" entity \textit{``Alaska Health Center"}. In terms of context augmentation, we randomly replace or insert some non-stop words (e.g., verb, adverb) in sentences to create new sentences as an example shown in Figure \ref{citation-figure2}. The augmented data does not change the meaning of original sentences but increase the diversity of the data. It can better help the model to learn contextual patterns and prevent the model focusing on specific PHI entities. Data augmentation is widely used in many NLP tasks \cite{xie2017data,ratner2017learning,kobayashi2018contextual,yu2018qanet,DBLP:conf/aclnut/BodapatiYA19,DBLP:conf/emnlp/WeiZ19} to improve model's robustness and generalizability. However, to the best of our knowledge, no work explores its potential in the clinical text de-identification task.

We test two LSTM-based models: NeuroNER \cite{Dernoncourt2017DeidentificationOP}, DeepAffix \cite{yadav-etal-2018-deep} and one BERT-based \cite{Devlin2019BERTPO} model: ClinicalBERT \cite{alsentzer-etal-2019-publicly} with our \model. Cross-dataset evaluations on i2b2 2006 dataset and i2b2 2014 dataset show that \model can boost the models' generalization performance up to 8.6\% in terms of F1-score. We also discuss how much augmentation we need and conduct the ablation study to explore the effect of PHI augmentation and context augmentation. To summarize, our \model is simple yet effective and can be used together with any existing machine learning-based de-identification systems to improve their generalizability on new datasets.

\section{\model}
To understand what factors lead to the poor generalization, we check some error examples and find that most of the PHI entities in these error examples do not appear in training set or appear as a different PHI type (e.g., Washington [Name v.s. Location]). We argue that neural models might focus on too much on specific entities (e.g., recognizing ``Washington" as ``Location") but fail to learn general language patterns (e.g., ``met" is not usually followed by a ``Location" entity but a ``Name" entity instead). Consequently, such unseen or Out-Of-Vocabulary PHI entities might be hard to be identified correctly, thus leading to lower performance. To help models better identify these unseen PHI entities, we may encourage models to learn contextual patterns or linguistic characteristics and prevent models focusing too much on specific PHI tokens. 

\begin{figure}[t]
    \centering
    \includegraphics[width=\linewidth]{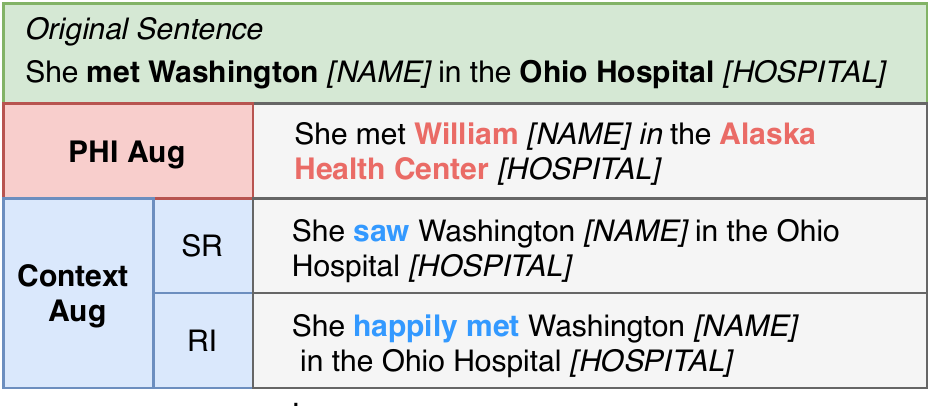}
    \vspace{-15pt}
    \caption{Toy examples of our \model data augmentation. SR: synonym replacement. RI: random insertion.}
    \label{citation-figure2}
    \vspace{-10pt}
\end{figure}

\noindent\textbf{PHI Augmentation.} To achieve this goal, we first introduce PHI augmentation: create more training corpora by replacing original PHI entities in the sentence with other named-entities of the same PHI type. For example, in Figure \ref{citation-figure2}, \textit{``Washington”} is replaced by a randomly-sampled Name entity \textit{``William”} and \textit{``Ohio Hospital”} is replaced by an randomly-sampled Hospital entity \textit{``Alaska Health Center”}. 

We construct 11 candidate lists for sampling different PHI types. The lists are either obtained by scraping the online web sources (e.g., Wikipedia Lists) or by randomly generating based on pre-defined regular expressions (the number and the source of each candidate list is shown in Table \ref{named-entity-list}).

\begin{table*}[t]
\centering
\resizebox{\linewidth}{!}{
\begin{tabular}{lclclc}
\hline\hline
\multicolumn{6}{l}{\textbf{Scraped from the Web}} \\ \hline
\multicolumn{1}{c|}{\textbf{PHI Type}} & \multicolumn{1}{c|}{\textbf{Number}} & \multicolumn{4}{c}{\textbf{Source}} \\ \hline
\multicolumn{1}{l|}{Organization} & \multicolumn{1}{c|}{1,300} & \multicolumn{4}{l}{\href{https://en.wikipedia.org/wiki/Category:Lists\_of\_organizations}{https://en.wikipedia.org/wiki/Category:Lists\_of\_organizations}} \\ \hline
\multicolumn{1}{c|}{\multirow{2}{*}{Hospital}} & \multicolumn{1}{c|}{\multirow{2}{*}{5,400}} & \multicolumn{4}{l}{\href{https://en.wikipedia.org/wiki/Lists\_of\_hospitals\_in\_the\_United\_States}{https://en.wikipedia.org/wiki/Lists\_of\_hospitals\_in\_the\_United\_States}} \\ 
\multicolumn{1}{l|}{} & \multicolumn{1}{c|}{} & \multicolumn{4}{l}{\href{https://www.hospitalsafetygrade.org/all-hospitals}{https://www.hospitalsafetygrade.org/all-hospitals}} \\ \hline
\multicolumn{1}{c|}{\multirow{3}{*}{Location}} & \multicolumn{1}{c|}{\multirow{3}{*}{27,500}} & \multicolumn{4}{l}{\href{https://en.wikipedia.org/wiki/List\_of\_Main\_Street\_Programs\_in\_the\_United\_States}{https://en.wikipedia.org/wiki/List\_of\_Main\_Street\_Programs\_in\_the\_United\_States}} \\ 
\multicolumn{1}{l|}{} & \multicolumn{1}{c|}{} & \multicolumn{4}{l}{\href{https://en.wikipedia.org/wiki/List\_of\_United\_States\_cities\_by\_area}{https://en.wikipedia.org/wiki/List\_of\_United\_States\_cities\_by\_area}} \\ 
\multicolumn{1}{l|}{} & \multicolumn{1}{c|}{} & \multicolumn{4}{l}{\href{https://en.wikipedia.org/wiki/List\_of\_United\_States\_cities\_by\_population}{https://en.wikipedia.org/wiki/List\_of\_United\_States\_cities\_by\_population}} \\ \hline
\multicolumn{1}{c|}{Patient} & \multicolumn{1}{c|}{14,900} & \multicolumn{4}{l}{\href{https://en.wikipedia.org/wiki/List\_of\_most\_popular\_given\_names}{https://en.wikipedia.org/wiki/List\_of\_most\_popular\_given\_names}} \\ \cline{1-2}
\multicolumn{1}{c|}{Doctor} & \multicolumn{1}{c|}{18,000} & \multicolumn{4}{l}{\href{https://en.wikipedia.org/wiki/List\_of\_most\_common\_surnames\_in\_North\_America}{https://en.wikipedia.org/wiki/List\_of\_most\_common\_surnames\_in\_North\_America}} \\ \hline\hline
\multicolumn{6}{l}{\textbf{Randomly Generated by Python scripts based on Regular Expressions}} \\ \hline
\multicolumn{1}{c|}{ID} & \multicolumn{1}{c|}{20,000} & \multicolumn{1}{l|}{\, \qquad Username \qquad \,} & \multicolumn{1}{c|}{\, \qquad 3,000 \qquad \,} & \multicolumn{1}{c|}{Zip} & 4,000 \\ \hline
\multicolumn{1}{c|}{Date} & \multicolumn{1}{c|}{32,900} & \multicolumn{1}{c|}{\, \qquad Phone \qquad \,} & \multicolumn{1}{c|}{\, \qquad 21,000 \qquad \,} & \multicolumn{1}{l|}{Medical Record} & 4,900 \\ \hline
\end{tabular}
}
\caption{\label{named-entity-list}The named-entity lists used for PHI augmentation, which are scraped from the Web or randomly generated.}
\end{table*}

\begin{table*}[t]
\centering
\resizebox{\linewidth}{!}{%
\begin{tabular}{l|l|c|c||l|c|c|c|c|c|c}
\hline
\multicolumn{2}{c|}{\multirow{2}{*}{}} & \multirow{2}{*}{\textbf{i2b2 2006}} & \multirow{2}{*}{\textbf{i2b2 2014}} & \multirow{2}{*}{\textbf{\begin{tabular}[c]{@{}l@{}}\# PHI of \\ each type\end{tabular}}} & \multicolumn{3}{c|}{\textbf{i2b2 2006}} & \multicolumn{3}{c}{\textbf{i2b2 2014}} \\ \cline{6-11} 
\multicolumn{2}{c|}{} &  &  &  & \textbf{Train} & \textbf{Dev} & \textbf{Test} & \textbf{Train} & \textbf{Dev} & \textbf{Test} \\ \hline
\multirow{4}{*}{\textbf{\#notes}} & \textbf{Train} & 622 & 912 & CONTACT & 159 & 32 & 41 & 394 & 31 & 96 \\ \cline{2-4}
 & \textbf{Dev} & 90 & 132 & DATE & 4887 & 649 & 1562 & 9102 & 974 & 2268 \\ \cline{2-4}
 & \textbf{Test} & 177 & 260 & ID & 3399 & 527 & 883 & 1000 & 166 & 312 \\ \cline{2-4}
 & \textbf{Total} & 889 & 1304 & LOCATION & 1761 & 252 & 648 & 3161 & 433 & 919 \\ \cline{1-4}
\multicolumn{2}{l|}{\textbf{\#avg tokens / note}} & 631.7 & 810.8 & NAME & 3163 & 452 & 1064 & 5156 & 745 & 1439 \\ \hline
\multicolumn{2}{c|}{\textbf{\#avg PHI / note}} & 21.9 & 20.1 & Total & 13369 & 1912 & 4198 & 18813 & 2349 & 5034 \\ \hline
\end{tabular}%
}
\caption{Statistics of the i2b2 2006 and 2014 datasets.}
\label{tbl:statistics}
\end{table*}

\noindent\textbf{Context Augmentation.} To further help models focus on contextual patterns and reduce overfitting, inspired by previous work \cite{DBLP:conf/emnlp/WeiZ19}, we leverage two text editing techniques: synonym replacement (SR) and random insertion (RI) to modify background context for data augmentation (examples are shown in Figure \ref{citation-figure2}). Specifically, SR is implemented by finding four types of non-stopping words (adjectives, verbs, adverbs and nouns) in sentences, and then replacing them with synonyms from WordNet \cite{Fellbaum1998WordNet}.
RI is implemented by inserting random adverbs in front of verbs and adjectives in sentences, as well as inserting random adjectives in front of nouns in sentences. 

For each sentence containing PHI entities in the corpus, we can apply both PHI augmentation and Context augmentation to obtain the augmented data $D_{aug}$. We can run $\alpha$ times (by setting different random seeds) to obtain different sizes of augmented data (e.g., $\alpha$ = 2 means augmenting the original dataset twice). Though with the $\alpha$ increases, we can obtain larger augmented training corpora, it may also bring much noise. We recommend a small value for $\alpha$ (See more discussions in Section \ref{sec:how_much_aug}). Then we merge the $D_{aug}$ with the original dataset $D$ to form the final dataset $D_{new}$ for training: $D_{new}$ = $D$ $\cup$ $\alpha$ $D_{aug}$.

In summary,  \model can significantly increase the diversity of training data without involving more labeling efforts. The augmented data can increase data diversity and enrich contextual patterns, which could prevent the model focusing too much on specific PHI entities and encourage it to learn general language patterns.

\section{Experimental Setup}
\subsection{Datasets}
We adopt two widely-used de-identification datasets: i2b2 2006 dataset and i2b2 2014 dataset, and split them into training, validation and testing set with proportion of 7:1:2, based on notes number. We remove low frequency (occur less than 20 times) PHI types from the datasets. To avoid PHI inconsistency between the two datasets, we map and merge some fine-grained level PHI types into a coarse-grained level type, and finally preserve five PHI categories: Name (Doctor, Patient, Username), Location (Hospital, Location, Zip, Organization), Date, ID (ID, Medical Record), Contact (Phone). The statistics of the datasets are shown in Table \ref{tbl:statistics}.

\subsection{Setup}
\noindent\textbf{Base Models.} We select two LSTM-based models: NeuroNER \citep{Dernoncourt2017DeidentificationOP}\footnote{\href{https://github.com/Franck-Dernoncourt/NeuroNER}{https://github.com/Franck-Dernoncourt/NeuroNER}}, DeepAffix \cite{yadav-etal-2018-deep}\footnote{\href{https://github.com/vikas95/Pref\_Suff\_Span\_NN}{https://github.com/vikas95/Pref\_Suff\_Span\_NN}} and one BERT model: ClinicalBERT \cite{alsentzer-etal-2019-publicly}\footnote{\href{https://github.com/EmilyAlsentzer/clinicalBERT}{https://github.com/EmilyAlsentzer/clinicalBERT}}. All hyperparameters are kept the same as the original papers.

\noindent\textbf{Evaluation.} To evaluate models' generalizability, we use the cross-dataset test on the two i2b2 challenge datasets: (1) Train the model on i2b2 2006 training set, and test on the whole i2b2 2014 dataset (Train + Dev + Test) (abbreviated as ``2006$\rightarrow$2014") (2) Train the model on i2b2 2014 training set, and test on the whole i2b2 2006 dataset (Train + Dev + Test) (abbreviated as ``2014$\rightarrow$2006"). 
For all experiments, we average results from five runs. We follow \citet{Dernoncourt2017DeidentificationOP} and report the micro-F1 score on binary token level.

\begin{table*}[t]
\centering
\begin{tabular}{l|ccccc}

\multicolumn{6}{c}{\textbf{Trained on i2b2 2006, Tested on i2b2 2014}}                                                                          \\ \hline
\multicolumn{1}{c|}{}                        & \multicolumn{5}{c}{Training Data Size}                                                     \\ \cline{2-6} 
\multicolumn{1}{c|}{\multirow{-2}{*}{Model}} & 20\%     & 40\%     & 60\%                            & 80\%                           & 100\%    \\ \hline
NeuroNER \cite{Dernoncourt2017DeidentificationOP}                             & 0.5990 & 0.6021 & 0.6364                        &  0.6436 & 0.6482 \\ 
 \quad + \textbf{\model}                                      & \textbf{0.6670} & \textbf{0.6979} & \textbf{0.7025}                        & \textbf{0.7063}                        & \textbf{0.7166}  \\ \hline 
DeepAffix \cite{yadav-etal-2018-deep}                             & 0.5590  & 0.5875 & 0.6069                        & 0.5976                        & 0.6118 \\ 
 \quad + \textbf{\model}                                      & \textbf{0.6699} & \textbf{0.6543} & \textbf{0.6905}                        & \textbf{0.7170}                         & \textbf{0.6982} \\ \hline
 ClinicalBERT \cite{alsentzer-etal-2019-publicly}  & 0.7055 & 0.7149 & 0.7351 & 0.7454 & 0.7519  \\ 
  \quad  + \textbf{\model}   & \textbf{0.7310} & \textbf{0.7412} & \textbf{0.7500}  & \textbf{0.7586} & \textbf{0.7569} \\ \hline
\multicolumn{6}{c}{}\\ 
\multicolumn{6}{c}{\textbf{Trained on i2b2 2014, Tested on i2b2 2006}}                                                                      \\ \hline
\multicolumn{1}{c|}{}                        & \multicolumn{5}{c}{Training Data Size}                                                     \\ \cline{2-6} 
\multicolumn{1}{c|}{\multirow{-2}{*}{Model}} & 20\%     & 40\%     & 60\%                            & 80\%                            & 100\%    \\ \hline
NeuroNER \cite{Dernoncourt2017DeidentificationOP}                                 & 0.7303 & 0.7513 & 0.7864                        & 0.7891                        & 0.7936 \\ 
 \quad + \textbf{\model}                                     & \textbf{0.7911} & \textbf{0.7944} & \textbf{0.8135}                        & \textbf{0.8175}                              & \textbf{0.8051} \\ \hline
DeepAffix \cite{yadav-etal-2018-deep}                               & 0.6950 & 0.7467 & 0.7852                        & 0.7774                        & 0.7736  \\ 
 \quad + \textbf{\model}                                      & \textbf{0.7523} & \textbf{0.7706} & \textbf{0.7919} & \textbf{0.7827}                        & \textbf{0.8085} \\ \hline
ClinicalBERT \cite{alsentzer-etal-2019-publicly}  & 0.8989 & 0.9043 & 0.9030 & 0.9069 & 0.9123  \\
  \quad  + \textbf{\model}   & \textbf{0.9004} & \textbf{0.9076} & \textbf{0.9059}  & \textbf{0.9078} & \textbf{0.9145} \\ \hline
\end{tabular}
\caption{\label{citation-table2}
Cross-dataset test performance (micro-F1 score on binary token level) on two experiment settings for models with and without \model on different training set sizes. All the numbers are the average from 5 runs.}
\end{table*}

\section{Results}
\subsection{Does \model improve generalization?}
In our preliminary experiments, we find that poor generalization tends to be more severe when the training set size is small. 
Thus, we consider the following training set fractions $(\%)$: $\left\{20,40,60,80,100 \right\}$ and we set the augmentation factor $\alpha$ = 2 considering both effectiveness and time-efficiency (See the influence of $\alpha$ in Section \ref{sec:how_much_aug}). 
Table \ref{citation-table2} shows the overall results, and interesting findings include:

(1) \model improves the generalizability of each de-identification model under different training sizes consistently. The results are not surprising as both PHI augmentation and context augmentation increase linguistic richness and enable models to focus more on language patterns, so as to help to train more generalized models.

(2) In general, the performance boost is large when the training data size is relatively small. This is because \model plays larger role at the low-resource case as it can significantly increase data diversity, language patterns, and linguistic richness. 

(3) The performance boost on the BERT-based model is less obvious than that on LSTM-based models. Since ClinicalBERT has already been pre-trained on large-scale corpus: MIMIC-III clinical notes \cite{Johnson2016MIMICIIIAF}. It is reasonable that the augmented data does not lead to large boost on ClinicalBERT. But there is still significant boost when training data size is relatively small.

(4) The boost on the setting ``2006$\rightarrow$2014" is larger than that in the setting ``2014$\rightarrow$2006".  Because i2b2 2014 dataset has more data and more comprehensive PHI patterns than i2b2 2006 dataset. Data augmentation is usually more effective when the training set size is smaller \cite{DBLP:conf/emnlp/WeiZ19}.

\begin{figure}[t]
    \centering
    \includegraphics[width=\linewidth]{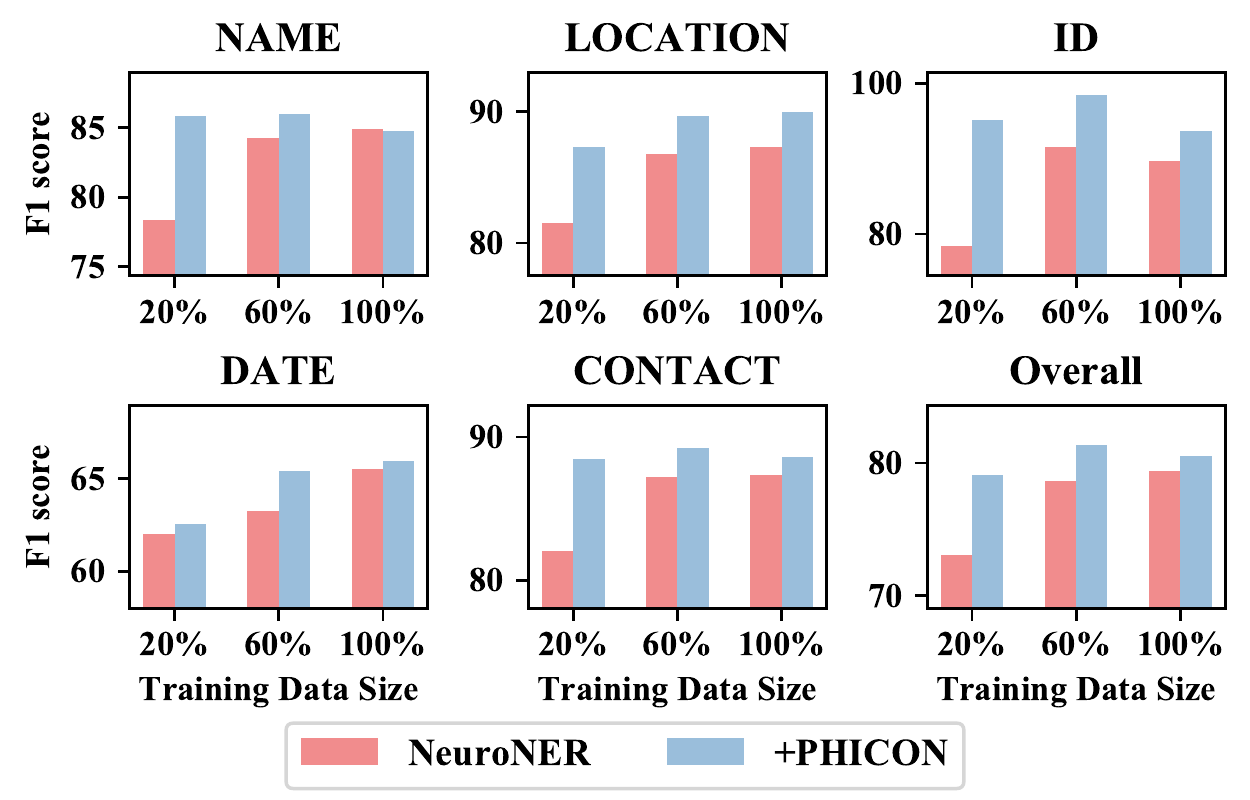}
    \caption{Performance of NeuroNER w/o and w/ \model on each PHI type (setting: 2014$\rightarrow$2006)}
    \label{fig:type_results}
\end{figure}

\begin{figure}[t]
  \centering
  \includegraphics[width=1.0\linewidth]{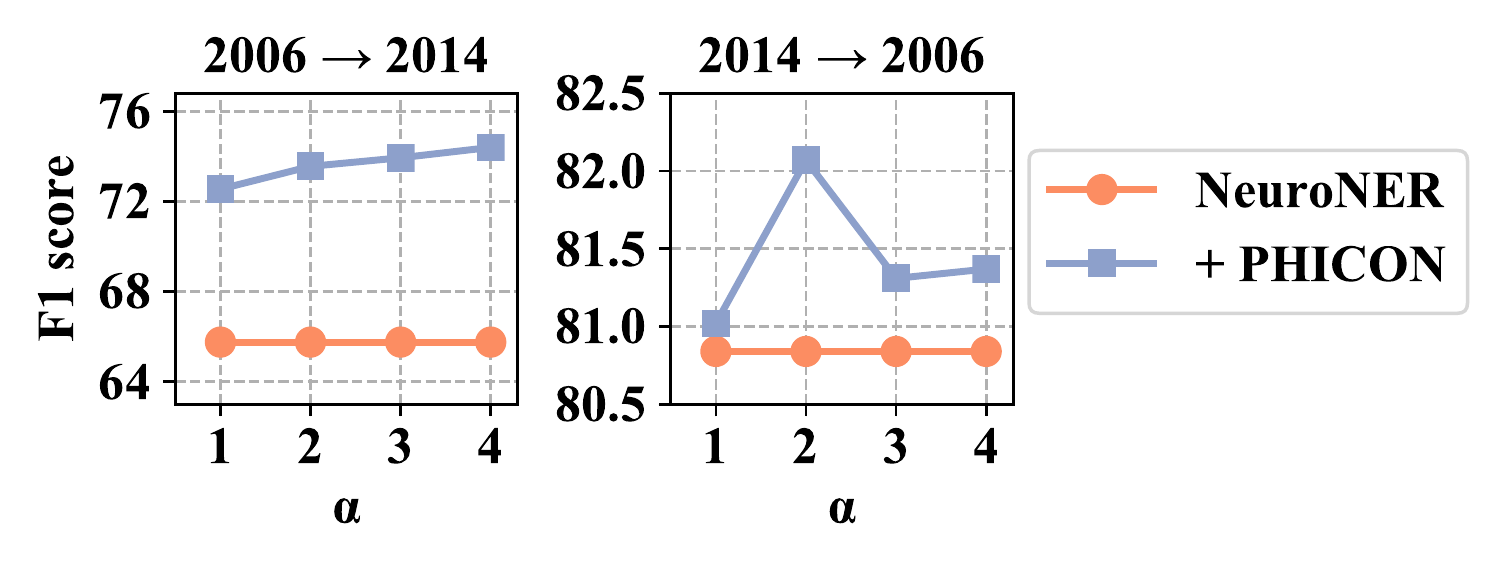}
  \caption{\label{fig_alpha_para}Data augmentation under different augmentation factors can boost model generalization. The left picture indicates that the model is trained on i2b2 2006 dataset and evaluated on i2b2 2014 validation set.}
\end{figure}

\vspace{2pt}
\noindent \textbf{Improvement for each PHI category}. To further understand \model, we show the performance ( ``2014$\rightarrow$2006") of the base model NeuroNER and NeuroNER + \model on each category of PHI in Figure \ref{fig:type_results}. Firstly, we can see that when the training data is relatively small (e.g., 20\%), the improvement on each PHI category is generally significant. With the training set size increases, the contribution of the augmented data becomes small. However, for the PHI categories that have less training data in the dataset (e.g., Location and ID; See Table \ref{tbl:statistics}), \model still contributes much improvement. Thus, we conclude that \model may be more helpful in the low-resource training data case.

\subsection{How much augmentation?}
\label{sec:how_much_aug}
In this section, we discuss the influence of the augmentation factor, $\alpha$, on the cross-dataset test performance. 
In Figure \ref{fig_alpha_para}, we report the performance on dev set based on the model NeuroNER for $\alpha$ = $\left\{1,2,3,4\right\}$. 
In the first setting (``2006$\rightarrow$2014"), we can see the performance is steadily boosted with the increase of the factor $\alpha$; while in the second setting (``2014$\rightarrow$2006"), the performance first goes up and then drops down. This difference might be caused by the data size of the two datasets (2014 dataset is larger). When the corpus is large, enlarging the augmentation factor might not lead to better performance, as the real data may have already covered very diverse language patterns. In addition, more augmented data might bring some noise, which could decrease the performance. In terms of time efficiency, when $\alpha$ is increased by 1, the training time would roughly double if we set the same epoch number. So considering effectiveness, efficiency and data size, we recommend to set $\alpha$ a relative small value (e.g., 2) in the real application.

\begin{table}[t]
\centering
\resizebox{0.95\linewidth}{!}{%
\begin{tabular}{lcc}
\hline
\textbf{Model} & \textbf{2006 $\rightarrow$ 2014} & \textbf{2014 $\rightarrow$ 2006} \\ \hline
NeuroNER     &  0.648    & 0.794     \\ 
+ PHI Aug      & 0.670     &  0.804    \\ 
+ Context Aug  &  0.659    &  0.803    \\ 
+ \textbf{\model} & \textbf{0.717}     &  \textbf{0.805}    \\ \hline
\end{tabular}%
}
\caption{\label{table-ablation-nalysis}Ablation study on \model. PHI augmentation and context augmentation contribute to the overall generalization boost.}
\end{table}

\subsection{Ablation Study}
In this section, we perform an ablation study on \model based on NeuroNER to explore the effect of each component: PHI augmentation and context augmentation. 
Table \ref{table-ablation-nalysis} shows that the two components of \model both contribute to boosting model generalization. Performance boost from PHI augmentation is obvious than context augmentation, i.e., PHI augmentation plays a major role. When combining both, \model results in larger boost than each of them.

\vspace{-3pt}
\section{Conclusion}
\vspace{-3pt}
In this paper, we explore the generalization issue on clinical text de-identification task. We propose a data augmentation method named \model that augments both PHI and context to boost model generalization. The augmented data can increase data diversity and enrich contextual patterns in training data, which may prevent the model overfitting on specific PHI entities and encourage it to focus more on language patterns. Experimental results demonstrate that our \model can help improve models' generalizability, especially in the low-resource training case (i.e., the size of the original training set is small). We also discuss how much augmentation to use and how each augmentation method influences the performance. In the future research, we will explore more advanced data augmentation techniques for improving the de-identification models' generalization performance.

\section*{Acknowledgments}
We thank Prof. Kwong-Sak LEUNG and Sunny Lai in The Chinese University of Hong Kong as well as anonymous reviewers for their helpful comments.


\end{document}